\documentclass{article}

\usepackage[preprint]{neurips_2024}

\usepackage[utf8]{inputenc} %
\usepackage[T1]{fontenc}    %
\usepackage{hyperref}       %
\usepackage{url}            %
\usepackage{booktabs}       %
\usepackage{amsfonts}       %
\usepackage{amsmath}
\usepackage{amssymb}
\usepackage{bm}
\usepackage{nicefrac}       %
\usepackage{microtype}      %
\usepackage{xcolor}         %
\usepackage{graphicx}
\usepackage{subcaption}
\usepackage{wrapfig}
\usepackage{multirow}
\usepackage{multicol}
\usepackage{tabularx}

\def\Figref#1{Figure~\ref{#1}}

\def\eqref#1{Eq.~(\ref{#1})}

\def\1{\bm{1}}

\DeclareMathAlphabet{\mathsfit}{\encodingdefault}{\sfdefault}{m}{sl}
\SetMathAlphabet{\mathsfit}{bold}{\encodingdefault}{\sfdefault}{bx}{n}

\def\gA{{\mathcal{A}}}

\def\gD{{\mathcal{D}}}
\def\gE{{\mathcal{E}}}

\def\gM{{\mathcal{M}}}

\def\gQ{{\mathcal{Q}}}
\def\gR{{\mathcal{R}}}

\def\gT{{\mathcal{T}}}

\def\gV{{\mathcal{V}}}

\usepackage[most]{tcolorbox}
\usepackage{listings}
\usepackage{float}

\tcbset{
  aibox/.style={
    width=474.18663pt,
    top=10pt,
    colback=white,
    colframe=black,
    colbacktitle=black,
    enhanced,
    center,
    attach boxed title to top left={yshift=-0.1in,xshift=0.15in},
    boxed title style={boxrule=0pt,colframe=white,},
  }
}
\newtcolorbox{AIbox}[2][]{aibox,title=#2,#1}

\newcommand{\ours}[0]{ITA}
\newcommand{\ourbench}[0]{Med-Critics}
\newcommand{\oursfull}[0]{\textbf{I}terative \textbf{T}ree \textbf{A}nalysis}

\title{Iterative Tree Analysis for Medical Critics}

\author{%
  Zenan Huang\thanks{Corresponding author.}
  \quad~~Mingwei Li$^{*}$
  \quad~~Zheng Zhou
  \quad~~Youxin Jiang \\
  Baichuan Inc. \\
  \texttt{\{huangzenan,limingwei\}@baichuan-inc.com}
}

\begin{document}

\maketitle

\begin{abstract}
Large Language Models (LLMs) have been widely adopted across various domains, yet their application in the medical field poses unique challenges, particularly concerning the generation of hallucinations. 
Hallucinations in open-ended long medical text manifest as misleading critical claims, which are difficult to verify due to two reasons. First, critical claims are often deeply entangled within the text and cannot be extracted based solely on surface-level presentation.
Second, verifying these claims is challenging because surface-level token-based retrieval often lacks precise or specific evidence, leaving the claims unverifiable without deeper mechanism-based analysis.
In this paper, we introduce a novel method termed \oursfull~(\textbf{\ours}) for medical critics. \ours~is designed to extract implicit claims from long medical texts and verify each claim through an iterative and adaptive tree-like reasoning process.
This process involves a combination of top-down task decomposition and bottom-up evidence consolidation, enabling precise verification of complex medical claims through detailed mechanism-level reasoning.
Our extensive experiments demonstrate that \ours~significantly outperforms previous methods in detecting factual inaccuracies in complex medical text verification tasks by 10\%.
Additionally, we will release a comprehensive test set to the public, aiming to foster further advancements in research within this domain.
\end{abstract}

\section{Introduction}

Large language models (LLMs) have shown remarkable proficiency across a wide range of tasks.
However, in high-stake domains like evidence-based medicine, they face critical challenges that limit their practical reliability \cite{djulbegovicProgressEvidencebasedMedicine2017}.
Although LLMs excel in standardized medical benchmarks \cite{noriMedpromptO1Exploration2024}, such evaluations frequently overlook the nuanced and complex nature of real-world clinical scenarios.
To address this gap, future assessments should extend beyond traditional multiple-choice formats, such as MedQA \cite{jinWhatDiseaseDoes2020}, and incorporate open-ended, long-form evaluations. 
These evaluations should rigorously test the models' ability to verify complex and nuanced information \cite{schmidgallAgentClinicMultimodalAgent2024,fanAIHospitalBenchmarking2024}. 
Such assessments should incorporate intricate and implicit medical knowledge, challenging LLMs to demonstrate deeper understanding and reasoning. 
Accurate verification of response is particularly crucial in medical applications, as LLMs frequently generate hallucinations—erroneous or fabricated outputs \cite{liHaluEvalLargeScaleHallucination2023,chengEvaluatingHallucinationsChinese2023}. 
By emphasizing detailed factuality verification, researchers can better understand LLMs' strengths and limitations, guiding their improvement and enhancing their reliability in critical applications.

The rapid development of LLM-based QA system and LLM-as-a-Judge pipelines \cite{zhengJudgingLLMJudgeMTBench2023,chenMLLMJudgeAssessingMultimodal2024} has produced numerous solutions for generating accurate answers and assertions.
However, factuality verification remains a significant challenge, particularly in domains like medicine.
Current methods \cite{minFActScoreFinegrainedAtomic2023} typically divide factuality verification into two tasks: (i) determining whether a claim is factually correct, and (ii) evaluating whether a claim is supported by evidence \cite{tranRARERetrievalAugmentedReasoning2024}.
These approaches often rely on surface-level input descriptions, which are insufficient for complex, long-form medical statements.

Medical factuality verification is especially challenging due to the intricate relationships among medical concepts, symptoms, and treatments etc., which often cannot be validated through simple queries.
This process requires understanding implicit causal effects and constructing detailed chains of evidence.
Existing automated methods, such as fact-checking \cite{liHaluEvalLargeScaleHallucination2023} and knowledge-based systems \cite{chenFELMBenchmarkingFactuality2023,vuFreshLLMsRefreshingLarge2023}, typically compare claims to static databases or predefined reference answers.
While effective for simple claims, these methods struggle with the complexity and interconnectedness of long-form medical statements, highlighting the need for more advanced investigative processes and dynamic evaluation systems.

In this paper, we introduce a novel framework, \oursfull~(\textbf{\ours}), designed to identify and rectify factual inaccuracies within medical claims articulated in natural language.
\ours~innovatively tackles the primary task by implementing a verification process that systematically manages specific sub-claims. 
This process involves recursively verifying sub-claims and constructing a verification tree. 
The tree structure is developed based on the sub-claims and the currently available information retrieved for each sub-tree. 
By consolidating the verification tree from the bottom up, our approach achieves complex verification objectives. 
This divide-and-conquer strategy simplifies the task into manageable sub-claims and, by integrating more reliable external references, facilitates the progressive exploration of challenging verification tasks. 
In summary, our contributions include:
\begin{itemize}
\item We introduce \textbf{\ours}, a novel system designed to enhance medical factuality verification through the use of adaptive tree-of-thoughts reasoning. This approach efficiently extracts atomic claims from the original text and constructs a tree of evidence to support true or false judgments, thereby improving the accuracy and reliability of claim verification.

\item \textbf{\ours} offers a unified and versatile framework for medical claim detection by spanning and consolidating sub-trees with retrieved external reference information. This capability allows for the comprehensive illustration and validation of distinct vital claims within the input query text, facilitating a more nuanced and detailed analysis of medical information.

\item To support the evaluation of the medical verification task, we curate a dataset called \textbf{\ourbench}, which includes a fine-grained checklist. This dataset facilitates a more explicit understanding of verification methods.
\end{itemize}

\section{Related Works}

\paragraph{Factuality verification.}
Factuality is a crucial attribute for LLMs, as controlling the generation of misleading information poses significant challenges. 
\citet{chengEvaluatingHallucinationsChinese2023} utilize GPT-4 to automatically evaluate whether a model's output is hallucinated.
In dialogue contexts, \citet{luoHalluDialLargeScaleBenchmark2024} develop HalluJudge, a specialized model for dialogue-level hallucination evaluation, based on a large-scale benchmark dataset. 
For long-form content, \citet{minFActScoreFinegrainedAtomic2023} propose FactScore, which assesses the factuality by breaking content into atomic sentences and retrieving relevant information from Wikipedia.
Enhancing this approach, \citet{weiLongformFactualityLarge2024} integrate a Google Search API to enable more flexible and robust factuality assessments.
For multi-modal content, \citet{jingFaithScoreFinegrainedEvaluations2024} introduce FaithScore, offering fine-grained reliability evaluations.
Despite these advancements, most existing methods focus on surface-level information and may fall short in addressing the complexity of real-world verification tasks.

\paragraph{Adaptive retrieval.}
Adaptive retrieval is a key criterion for addressing the challenges of external document search, focusing on when and how extensively to retrieve information.
\citet{shaoEnhancingRetrievalAugmentedLarge2023} utilize iterative retrieval to generate more reliable outputs, an approach echoed by \citet{trivediInterleavingRetrievalChainThought2023}, who refine external knowledge iteratively to consolidate final answers.
Similarly, \citet{jiangRetrieveSummarizePlan2024} enhance the completeness of QA by progressively improving retrieved information.
\citet{jeongAdaptiveRAGLearningAdapt2024} propose adaptive retrieval techniques that train a smaller language model to categorize queries into no retrieval, single-step retrieval or multi-step retrieval.
\citet{baekProbingRAGSelfProbingGuide2024} rely on the LLMs’s internal knowledge to determine when retrieval is necessary. 
\citet{dingRetrieveOnlyWhen2024} and \citet{zhangRetrievalQAAssessingAdaptive2024} introduce the internal hallucination checks, triggering retrieval only when there is a risk of hallucination.
\citet{jiangActiveRetrievalAugmented2023} assess the necessity of retrieval based on next-token confidence.
Further advancements include \citet{yueInferenceScalingLongContext2024}, who iteratively insert RAG QA exemplars to improve RAG utilization, and \citet{xuSearchChainInteractivelyEnhancing2024} integrate a chain-of-query mechanism to dynamically expand knowledge verification from the initial claim.

\paragraph{Knowledge verification.}
Conflicting perspectives are inherent part in any field. 
While some conflicts can be resolved through source verification, others remain subjects of ongoing debate. 
LLMs often struggle to provide consistent outputs when presented with conflicting documents \citep{suDRAGINDynamicRetrieval2024}. 
LLMs typically prioritize coherence and persuasive narratives,  which can lead to biased or incomplete conclusions. To address this, it is crucial to guide LLMs toward a formal reasoning framework that relies on internal logic and external references \cite{waddenSciFactOpenOpendomainScientific2022,weiLargerLanguageModels2023}. 
Building on prior findings, \citet{bayatFactBenchDynamicBenchmark2024} propose categorizing information into three categories--\texttt{supported}, \texttt{unsupported}, and \texttt{undecidable}--by introducing an undecidable placeholder for cases where conclusive evidence is unavailable.

\section{VITAL: Verification and Iterative Tree Analysis}

Accurately evaluating the claims of long-form medical statements presents significant challenges due to the complexity of medical discourse, the interconnection of concepts, and the limitations of existing evaluation methods.
Current approaches often rely on surface-level assessments or static databases, which fail to capture the nuanced relationships among medical concepts, symptoms, and treatments \textit{etc}.
First, we advocate for a fine-grained evaluation approach that focuses on individual medical concepts and evidential chains than entire statements or sentences.
This granular method enables precise verification by independently assessing each concept and its supporting evidence, ensuring the evaluation reflects the intricacies of medical knowledge.
Second, we propose leveraging dynamic, up-to-date medical databases and reference searches.
By integrating these insights, we can enhance the reliability and accuracy of long-form medical text verification, ultimately improving the quality of medical applications.

\begin{figure}
    \centering
    \includegraphics[width=0.95\linewidth]{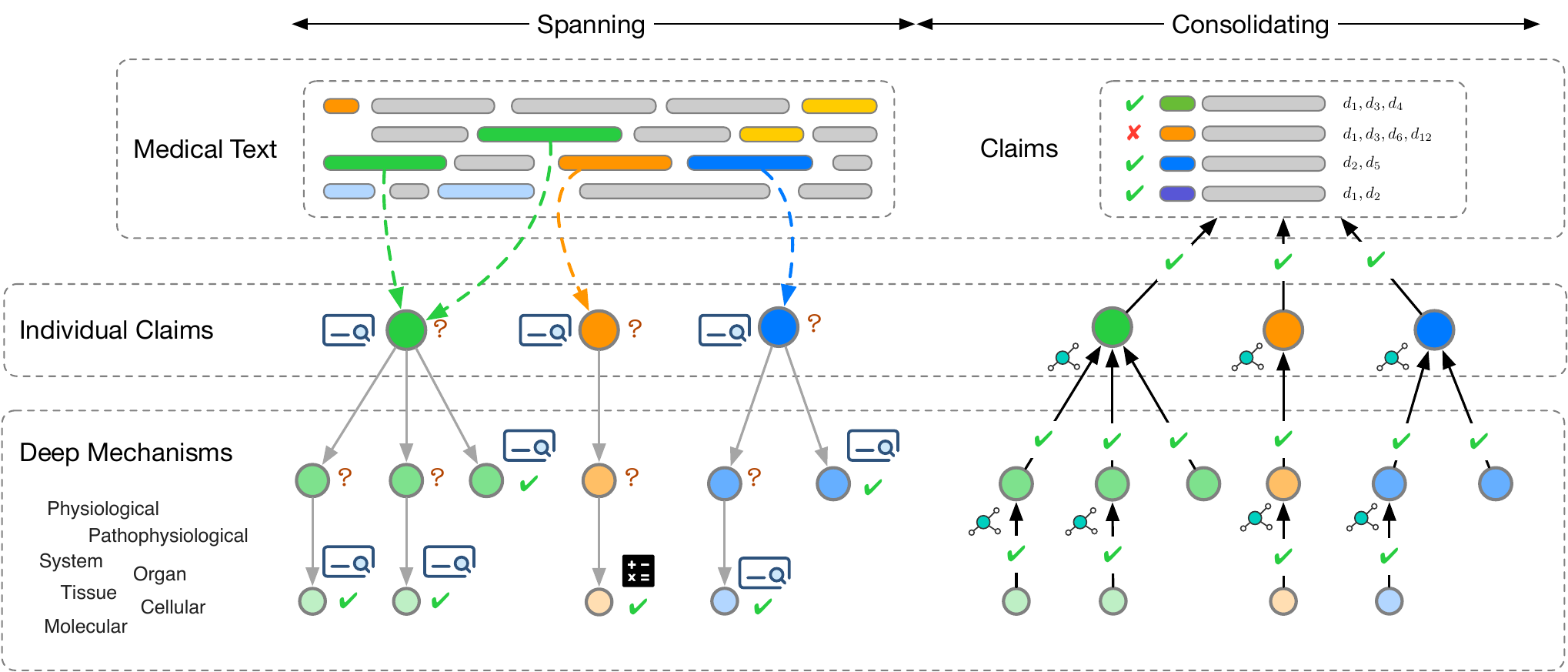}
    \caption{Overview of the \textbf{\ours} framework. During the tree-\textbf{spanning} stage, individual claims are extracted from the medical text, ensuring each claim is self-contained by incorporating information from other parts of the text. Verification sub-tasks are distributed recursively down to the leaf nodes. In the \textbf{consolidating} stage, deeper knowledge insights for each individual claim are used to determine whether to accept or reject its parent claim. The framework outputs the verification status of each individual claim along with its supporting references.}
    \label{fig:framework}
\end{figure}

As shown in \Figref{fig:framework}, the verification process involves extracting all verifiable claims from the input query and constructing a verification tree, $\gT$, based on these claims.
Medical claims under scrutiny often require complex reasoning chains that depend on fine-grained knowledge points to identify distinct claims accurately.
In some cases, the verification process demands indicator computation or the consolidation of multi-level claims, particularly when the combination of medical concepts is absent from external references or challenging to retrieve.
In such scenarios, it is crucial to gather sufficient evidence from external sources, such as recent research articles, to construct the implicit evidence chain necessary to determine whether the claim is supported.

\subsection{Problem Formalization and Overview}

We introduce the following components of \ours: the large language model $\gM$, claim variables $\gV := \{v_{i}\}$, evidence relations $\mathcal{E} := \{(v_{i}, v_{j})\}$, verification tree $\gT := \{\gV, \gE\}$, input query $q$, information source $\gD$, and the retriever $\gR$.
The verification task $\gQ$ involves determining the factuality of a statement by constructing a logical sequence of the thoughts supported by external references. 
Given an input query $q$, our objective is to construct an optimal thought tree $\gT$ that adaptively generate verification sub-trees and retrieves relevant document sources $\gD$ as evidence.
Specifically, for a given $q$, the LLM $\gM$ is prompted to iteratively expand the thought tree $\gT$, ensuring each node and relation in tree is grounded in retrieved evidence:
\begin{equation}\label{eq:overview}
    \gA = f_{\text{verify}}(q;\gM,\gR,\gD,\gT),
\end{equation}
where $\gA$ is a set of verified claims $a_{i}$, each annotated with a judgment flag \textsc{State} and \textsc{Reason}.
The ultimate objective is to construct logical sequences that determine whether all claims in a query can be substantiated by evidence, adhering to specific references. 
These references encompass diverse sources, including Wikipedia, textbooks, expert explanations, and precise calculator.
By embracing this broader definition, we establish a unified framework for addressing issues of factual accuracy.

While input content can be readily divided into sentences and short representations, identifying precise atomic claims requires a higher level of granularity. 
Often, information not immediately apparent is crucial for substantiating these claims. 
For instance, some claims necessitate multi-hop verification, while others may be susceptible to spurious relationships, requiring careful scrutiny to avoid misleading confounders.
In the context of a tree-based representation, these claims can be structured as a graph, with each node symbolizing an minimum claim and the edges represent the relationships between claims.
Naively extracted claims often lack sufficient reasoning details for proper validation.
By systematically refining claims with the support of external fact-checking tools and rigorously analyzing the entire tree, we can ensure both the accuracy of individual claims and the integrity of the expanded tree.

\subsection{Verification with spanning tree}
The ultimate concept is exploration when the answer is not sure and consolidation when external information is enough.
After iteratively referring to external resources, the claims extracted from the initial query can be verified with an in-depth reasoning chain.

\paragraph{Spanning.}
The process of generating a verification tree begins by identifying critical claims from the initial statement, which serves as the foundation for both exploration and exploitation.
Depending on the specific medical problem, a claim can take various forms, such as a few words (\textit{e.g.}, the name of a drug), a line of an equation (\textit{e.g.}, a biomedical indicator), or a sentence expressing a cause-effect relationship (\textit{e.g.}, symptom inference).

Typically, a sub-claim should be ``atomic'' enough to enable language models to generate effective queries. 
Given a query containing task-specific information, the language model $\gM_{\text{generate}}$ is then prompted to produce the claims requiring verification:
\begin{equation}\label{eq:init_claims}
    \textsc{SubTree}(q,\textsc{Child}(q)) \leftarrow \gM_{\text{generate}}(q), \quad \textsc{Child}(q) = \{v'_{1}, v'_{2}, \dots\}.
\end{equation}
Here, $v'_{i}=\{\textsc{Claim},\textsc{State}\}$ is initially set to ``verifying''.
To ensure comprehensive factual checking, we follow the guidelines established by \citet{minFActScoreFinegrainedAtomic2023}, revising the proposed claims to be self-contained.
Once the node $v'_{i}$ is verified, it will be supplemented with reference information, a judgment reason, and an updated \textsc{State}. 
This updated node is denoted as $v^{*}_{i}=\{\textsc{Claim},\textsc{State},\textsc{Reason},\textsc{Ref}\}$.

Once the initial claims are generated, the verification tree $\gT$ is constructed by iteratively expanding the tree with new claims based on the information retrieved through the retriever $\gR$:
\begin{equation}\label{eq:spanning_subtree}
    \textsc{SubTree}(v'_{\text{cur}}, \textsc{Child}(v'_{\text{cur}})) \leftarrow \gM_{\text{span}}(v'_{\text{cur}}, \gR(v'_{\text{cur}}), \delta), \quad \delta = \textsc{Verify}(\gR(v'_{\text{cur}}), v'_{\text{cur}}).
\end{equation}
Here, $v'_{\text{cur}}$ is the root node of the current \textsc{SubTree}, and $\delta$ indicates the flag for spanning termination (\textit{e.g.}, $\delta:=$ \texttt{accept}/\texttt{reject} for termination, $\delta:=$ \texttt{unsubstantiated} for spanning). 
Once $\gM$ determines that further expansion is unnecessary or the maximum exploration condition is reached, the generated \textsc{SubTree} will contain only the node $v'_{\text{cur}}$. 
This will then trigger the bottom-up node verification consolidation process.

We propose a spanning method that involves generating follow-up queries based on a sub-tree structure originating from the root node. 
The root node contains meta-information, such as the abstract of the input theme, and decomposes extracted entities into leaf nodes. 
Each node is evaluated to maintain and update the current meta-information, facilitating a thorough verification process.

\paragraph{Consolidation.}
The tree consolidation process proceeds in a bottom-up manner from the leaf nodes, gradually integrating verified claims with the aid of external information.
Each spanning \textsc{SubTree} corresponds to a parent claim, accompanied by retrieved documents and associated child claims.
The state evaluator, $\gM_{\text{cons}}$, assesses progress toward solving the problem and serves as a claim consolidation mechanism. 
During the consolidation step, RAG-assisted verification begins at the leaf nodes, passing the processed claim's \textsc{State}—indicating whether the claim is supported—and the \textsc{Reason}, which provides the judgment, to the parent node.
This process modifies the parent node using reasoning results from its children:
\begin{equation}\label{eq:consolidating}
v^{*}_{\text{cur}} \leftarrow \gM_{\text{cons}}(v'_{\text{cur}},\{v^{*}_{1},v^{*}_{2},\dots\})~|~v^{*}_{i}\in \textsc{Child}(v'_{\text{cur}}),
\end{equation}
where $v'_{\text{cur}}$ is the root node of the current \textsc{SubTree} to be verified. 
A consolidation prompt reasons about the sub-tree to generate a scalar value for \textsc{State} (\textit{e.g.}, a score from 1 to 10) or a confidence level category (\textit{e.g.}, accept, reject, or unsubstantiated), which can be mapped to a numerical value. 
The specific categories may vary depending on the problem or the reasoning steps involved.

\begin{wrapfigure}{r}{0.5\textwidth}
    \vspace{-.4cm}
    \centering
    \includegraphics[width=\linewidth]{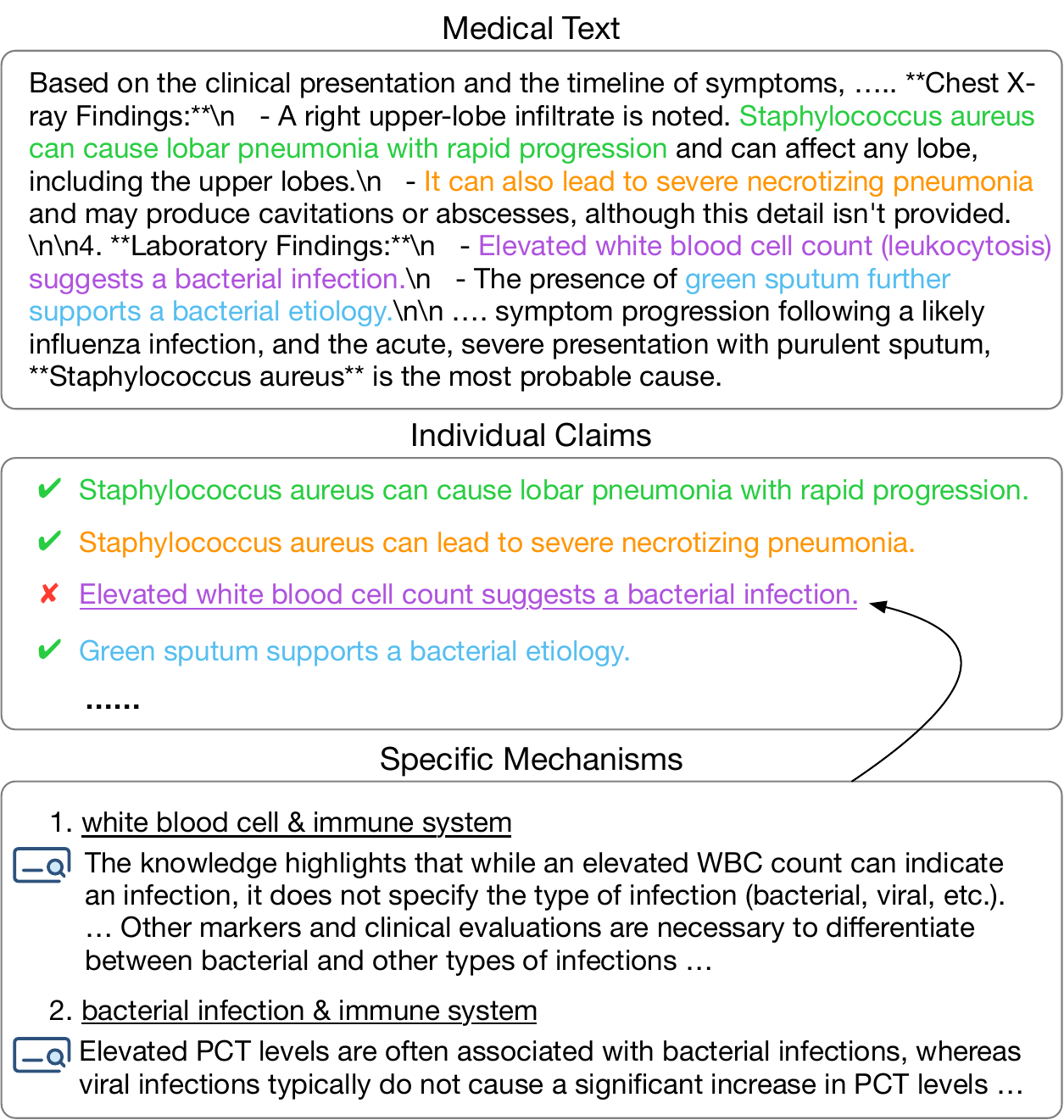}
    \caption{An example of verification}
    \label{fig:case_illustration}
    \vspace{-.8cm}
\end{wrapfigure}

Intuitively, when $v'_{\text{cur}}$ refers to symptoms and medication usage, $\textsc{Child}(v'_{\text{cur}})$ maintains information pieces such as organs, tissues, and medication treatments. 
By delving into the \textsc{Child} of the \textsc{Child}, we can explore relationships between cellular, protein, and molecular mechanisms, as illustrated in \Figref{fig:framework}.

\subsection{Individual Claims \& Retrieval}

Medical texts often contain critical claims embedded within the main narrative, much like intricate patterns woven into a complex textile. 
The surface presentation of these texts does not always transparently convey the underlying logical structure. 
To accurately identify and interpret key claims, we design a multi-hop claim refinement agent and distill its reasoning trace into LLM with fine-tuning.
This process allows for the extraction of significant claims that may otherwise remain obscured within the broader context.
As illustrated in \Figref{fig:case_illustration}, long-form text often contains numerous pieces of information that can be both mixed and ambiguous. 
A critical component of successful verification is ensuring that each \textsc{SubTree} is self-contained. 
This requires that each \textsc{SubTree} includes explicit entity names rather than pronouns or vague references, which could impede the verification process. 
We construct self-consistent factual statements from the original text to facilitate a more fine-grained evaluation, as depicted in \Figref{fig:case_illustration}. 
Each factual claim is a self-contained sentence that conveys information about diseases, conditions, medications, treatments, therapies, diagnoses, indicators, and side effects.

\paragraph{Query Generation.} 
For a given claim $v_{i}$, the retriever $\gR$, with the assistance of the LLM $\gM_{\text{query}}$, identifies the appropriate external sources to consult. 
The retrieval space of $\gR$ initially comprises a set of pre-built tools $\gR=\{R_{1}, R_{2}, \dots\}$, where each tool $R_{i}$ is tailored to retrieve specific types of information or to invoke an external calculator for evaluating indicator results. 
The selection process is guided by the following equation:
\begin{equation}\label{eq:query_generation}
    \{R_{i}, s\} \sim \gM_{\text{query}}(v'_{i}, \textsc{Parent}(v'_{i})).
\end{equation}
In this context, $R_{i}$ is determined using the meta-claim of its parent node as context, while $s$ represents the generated search query or coding task. 
This facilitates the distributed verification of claims $v'_{i}$, such as specific biomedical indicator computations that are contingent upon the parent node's meta-claim. 
Furthermore, certain molecular mechanisms related to the parent node's meta-claim may necessitate external information updates for accurate verification.

\paragraph{Retrieval and Selection.} 
Once information is retrieved, the retriever $\gR$ processes it to select relevant documents and extract key information. 
The external sources utilized can range from general search engines and medical textbooks to specialized calculators. 
Despite the diversity of these sources, the retrieval process consistently aims to provide evidence that supports the claims. 
Consequently, the retrieved information is re-ranked and selected based on its relevance to the claims:
\begin{equation}\label{eq:retrieval_selection}
    D \leftarrow \textsc{Rerank}(R_{i}(\epsilon)),\quad D = \{d_{1}, d_{2}, \dots\}.
\end{equation}
Here, $D$ represents the set of retrieved information, with each $d_{i}$ denoting the preprocessed content extracted from the documents. 
The \textsc{Rerank} function can be either rule-based or learned. 
For instance, it can be configured to prioritize scientific resources over less reliable sources, such as advertising. 
This curated information is essential for verifying claims and providing robust evidence for the consolidation process.

\section{Experiments}

In this section, we present the results of our end-to-end factuality evaluation. 
Long-form factuality assessment is challenging due to the difficulty in defining a definitive set of facts. 
To address this, we developed a fine-grained benchmark with specific claim modifications to evaluate the performance of our method, \ours. 
We also assessed the reliability of baseline verification methods and human annotators in processing long-form medical texts, supported by extensive statistical analyses.

\subsection{\ourbench~Curation}

Inspired by previous works \cite{scireTruthMirageEndtoEnd2024,nessMedFuzzExploringRobustness2024}, VITAL-Med is constructed by systematically extracting claims from medical texts, falsifying a subset of these claims, and generating paired factual and non-factual texts. 
For this purpose, We utilize a subset of the \textbf{MedQuAD} dataset \citep{benabachaQuestionentailmentApproachQuestion2019}, focusing on 20-30 sentence segments from medical texts.
MedQuAD, sourced from various National Institutes of Health websites, contains real-world medical QA pairs covering 37 question types, addressing topics such as treatments, diagnoses, and side effects.
The construction of \ourbench~involves the following steps:
\begin{enumerate}
\item \textbf{Claim Extraction:} Given a passage from a medical text, a LLM extracts a list of atomic claims breaking the content into its fundamental factual components.

\item \textbf{Claim Falsification:} To introduce controlled inaccuracies, one of the extracted claims is deliberately falsified. 
This involves adding a random error, such as a misleading viewpoint or distortion of critical information, to simulate realistic misinformation scenarios. 

\item \textbf{Text Paraphrasing:} Using the two sets of claims (factual and falsified), the LLM generates a paraphrase of the original text that maintains the factual integrity of the claims.

\item \textbf{Alternative Text Generation:} An alternative version of the text is created by incorporating the falsified claim, resulting in a non-factual narrative for evaluation purposes.
\end{enumerate}
This pipeline supports the creation of a comprehensive dataset aimed at evaluating models' ability to distinguish between factual and non-factual information within medical texts.
To enable the benchmark to effectively identify factual inaccuracies across multiple dimensions, we classify the test data into 6 primary categories:
\begin{itemize}
\item \textbf{Pathophysiology}: Covers the biological and physiological processes underlying diseases or injuries, providing a framework for understanding disease mechanisms.

\item \textbf{Medication}: Focuses on pharmacological treatments, including drug interactions, side effects, and therapeutic efficacy.

\item \textbf{Diagnosis}: Involves the identification and classification of diseases, with an emphasis on diagnostic criteria and methodologies.

\item \textbf{Symptom}: Describes the clinical manifestations of diseases, detailing symptoms and their relevance to specific medical conditions.

\item \textbf{Treatment}: Addresses therapeutic interventions, both medical and surgical, aimed at managing or curing diseases.

\item \textbf{Prevention}: Explores strategies and measures to prevent disease onset or recurrence, including lifestyle modifications and prophylactic treatments.
\end{itemize}

\begin{table}[ht]
\centering
\caption{The statistics of \ourbench~benchmark. The average length is measured based on the text, while the positive rate is defined by the proportion of claims that are factually correct.}
\label{tab:dataset-statistics}
\resizebox{\textwidth}{!}{%
\begin{tabular}{@{}lccccccc@{}}
\toprule
\multirow{2}{*}{Statistics} & \multicolumn{6}{c}{Open-Ended QA}                                                               & \multirow{2}{*}{Avg.} \\ \cmidrule(lr){2-7}
                            & pathophy. & medication & diagnosis & symptom & treatment & \multicolumn{1}{l}{prevention} &                       \\ \midrule
Num. Texts & 330   & 39    & 86    & 334   & 143   & 48    & 163.3 \\
Num. Claims  & 1435  & 65    & 195   & 4066  & 426   & 127   & 1052.3 \\
Avg. Tokens  & 243.2 & 204.6 & 232.7 & 514.7 & 198.3 & 217.4 & 268.5 \\
Postive Rate & 22.3\%  & 60.0\%  & 44.1\%  & 8.1\%  &  33.5\%   & 37.8\%  & 34.3\% \\ \bottomrule
\end{tabular}%
}
\end{table}

\subsection{Evaluation setups}
We concur with the notion that long-form input text should be evaluated at the granularity of individual facts \cite{minFActScoreFinegrainedAtomic2023}.
To this end, we adopt a fine-grained evaluation approach to assess the factuality of long-form medical texts.
This approach involves evaluating the factuality of each individual fact within the input text and reporting the overall performance of the method on the dataset.

\paragraph{Baselines.}
We evaluate the performance of our proposed method, \ours, against several state-of-the art factual consistency evaluation baselines. FActScore~\cite{minFActScoreFinegrainedAtomic2023} breaks texts into a series of atomic facts and then assigns a binary label to each fact individually.
FELM~\cite{chenFELMBenchmarkingFactuality2023a} segments the text into granular textual spans and evaluates the factual consistency of all spans collectively. 
RefChecker~\cite{huRefCheckerReferencebasedFinegrained2024} extracts knowledge triplets from the text and evaluates each triplet independently. 
LongFact~\cite{weiLongformFactualityLarge2024}, integrate a Google Search API to enable more flexible and iterative retrieval for assessing atomic facts.

\paragraph{Model performance.}
We also evaluate the performance of standard baseline LLMs in terms of their reliability accuracy when processing long-form medical texts. 
Our evaluation encompasses experiments conducted on subsets of \ourbench~open-ended medical QA challenges. 
These datasets are specifically designed to test the models’ ability to process complex medical information, handle nuanced reasoning, and generate accurate, contextually relevant responses.

\paragraph{Metrics.}
To evaluate our method's performance in assessing the factuality of long-form medical texts, we employ several key metrics. 
The accuracy measures the discrepancy between the ground truth and the factual verifications, providing a basic accuracy assessment.
We also use the $F_{1}@K$ metric \cite{weiLongformFactualityLarge2024}, which evaluates both precision and recall. 
This metric offers a comprehensive view of the model's factual accuracy.

\subsection{Main Results}

To highlight the faithfulness of our proposed model, \ours, we focus on two critical aspects: accurately identifying all fault claims and correctly evaluating each claim. 
However, the inherent unpredictability of generative models, combined with the absence of definitive rules for determining the number and nature of claims, makes this task particularly challenging. 
This ambiguity complicates the assessment of whether a judging system operates fairly. 
To address this, we utilize the checklist score with LLM-as-a-Judge to systematically evaluate claims.
Additionally, we tackle these challenges by introducing the \ourbench~benchmark, which provides a diverse set of predefined faulty claims with varying quantities for rigorious evaluation.

\begin{table}[ht]
\centering
\caption{Performance comparison of factual verification on the \ourbench}
\label{tab:methods-performance-comparison}
\resizebox{\textwidth}{!}{%
\begin{tabular}{@{}c|c|ccccccc@{}}
\toprule
\multirow{2}{*}{Method} &
  \multirow{2}{*}{Correct.} &
  \multicolumn{6}{c}{Falsified samples} &
  \multirow{2}{*}{Avg.} \\ \cmidrule(lr){3-8}
 &
   &
  pathophy. &
  medication &
  diagnosis &
  symptom &
  \multicolumn{1}{l}{treatment} &
  prevention &
   \\ \midrule
FActScore (GPT-3.5) & 70.1 & 65.8 & 63.1 & 60.6 & 69.4 & 67.3 & 81.9 & 68.3 \\
FActScore (GPT-4)   & 73.2 & 67.3 & 65.2 & 63.5 & 70.1 & 69.2 & 82.3 & 70.1 \\
FELM                & 69.7 & 50.8 & 60.3 & 60.6 & 70.5 & 73.2 & 80.4 & 66.5 \\
RefChecker          & 78.4 & 73.6 & 66.7 & 65.2 & 74.3 & 74.4 & 85.2 & 73.9 \\
Long-Fact           & 90.3 & 77.4 & 71.2 & 69.5 & 77.1 & 73.2 & 86.5 & 77.8 \\
\ours               & \textbf{93.4} & \textbf{88.9} & \textbf{84.2} & \textbf{78.4} & \textbf{86.1} & \textbf{87.9} & \textbf{95.8} & \textbf{87.8} \\ \bottomrule
\end{tabular}%
}
\end{table}

\paragraph{\ours~consistency with predefined factual ground-truth.}
In Table~\ref{tab:methods-performance-comparison}, we report the factual verification accuracy across different categories of facts.
The results demonstrate that \ours~achieves the highest accuracy across all categories.
The key performance gaps can be attributed to two primary factors: \ours’ ability to extract self-contained claims and its capacity to verify these claims through comprehensive analysis.
Baseline methods often focus on breaking claims into atomic units for verification. However, the results show that many failures arise from a lack of sufficient contextual information. In the medical domain, facts taken out of context may appear accurate when they are not. \ours’ key strength lies in striking a balance between contextual grounding and atomicity. During claim verification, \ours~emphasizes the context of a complete \textsc{SubTree}, enabling a balanced granularity and deeper analysis of external scientific knowledge.
This approach allows the evaluator to determine whether any sub-claim is false and assess whether the original claim can still be supported.
By maintaining this balance, \ours~significantly improves the overall performance in factual verification.

\begin{wraptable}{r}{0.5\textwidth} %
    \centering
    \caption{Claim extraction ablation}
\centering
\label{tab:fact_extraction_ablation}
\begin{tabular}{@{}ccccc@{}}
\toprule
Extract Method       & \multicolumn{1}{l}{Precision} & Recall & F1   \\ \midrule
\verb|ATOMIC|        & 74.3                          & 72.2   & 73.2 \\
\verb|DECONTEXT|     & 82.1                          & 80.7   & 81.4 \\
\verb|MED-DECONTEXT| & 88.2                          & 87.6   & 87.8 \\
\ours                & \textbf{89.4 }                         & \textbf{87.9}   & \textbf{88.6} \\ \bottomrule
\end{tabular}

\end{wraptable}

\paragraph{The impacts of claims extraction.}
To rigorously evaluate the influence of initial claim extraction on the \ours~verification process, we conducted a series of controlled experiments. 
In these experiments, we fixed the extracted claims while rerunning the \ours~subtree verification.
This setup allowed us to isolate and analyze the specific impact of claim extraction on the overall performance and accuracy of the system.
As shown in Table~\ref{tab:fact_extraction_ablation}, we tested three baseline claim extraction methods:
\verb|ATOMIC|: Extracts atomic claims based on the method proposed in \cite{minFActScoreFinegrainedAtomic2023}.
\verb|DECONTEXT|: Extends \verb|ATOMIC| by applying a decontextualization operation, prompting the model with each fact and its associated context for improved disambiguation \cite{weiLargerLanguageModels2023}.
\verb|MED-DECONTEXT|: Builds on \verb|DECONTEXT| but adapts the LLM prompts to emphasize medical information completion for domain-specific optimization.
The results indicate that the fine-tuned claim extractor in \ours~consistently outperforms these baselines when verifying target medical texts.

\begin{wrapfigure}{l}{0.3\textwidth} %
    \centering
    \vspace{-1em}
    \includegraphics[width=0.96\linewidth]{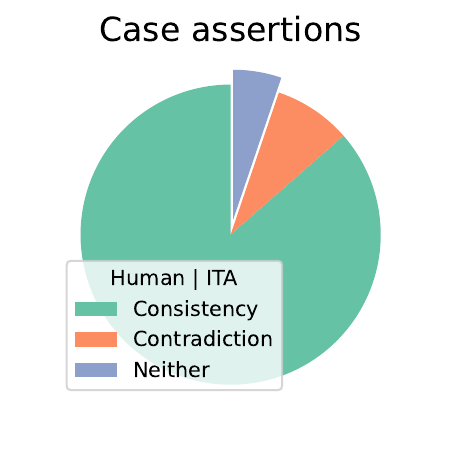}
    \vspace{-1em}
    \caption{Human evaluation}
    \label{fig:disagreement}
    \vspace{-1em}
\end{wrapfigure}
\paragraph{ITA consistency with human evaluation.}
To assess the reliability of \ours~in processing long-form medical texts, we conducted a human evaluation study involving three medical experts. 
The experts evaluated the reliability of claims verification labeled by \ours~on the \ourbench~benchmark. 
A random sample of 100 claims, categorized as either ``accept'' or ``reject'' by \ours, was selected for evaluation. 
Each expert assessed the factuality of the claims and was permitted to conduct internet searches to verify their accuracy.
As shown in Figure~\ref{fig:disagreement}, the results indicate a high level of agreement between the experts’ evaluations and the \ours~verification records.
While some claims were not fully verifiable due to the vague nature of the medical information or the lack of consensus within the academic community, unsubstantiated claims were minimal. 
These findings underscore the robustness of \ours~in accurately evaluating factuality in complex medical texts.

\begin{table}[ht]
\centering
\caption{Performance comparsion of facts verification on the \ourbench~open-ended QA, Accept/Reject/Unsubstantiated are count in fact level, where the precision is average precision of all samples, and $\text{Reall}{@5}$ and $\text{Reall}{@10}$ are the recall at top-5 and top-10.}
\label{tab:models-performance-comparsion}
\begin{tabular}{@{}cccc|c|cc@{}}
\toprule
\multirow{2}{*}{Model} & \multicolumn{6}{c}{Long-form reply}                                   \\ \cmidrule(l){2-7} 
                       & Accept    & Reject  & Unsubstantiated & Precision & $\text{Reall}{@5}$ & $\text{Reall}{@10}$ \\ \midrule
GPT-3.5                & 1314 & 223 & 2 & 85.73     & 84.69              & 56.33               \\
GPT-4o                  & 2264 & 202 & 5 & 90.44     & 95.71              & 83.17               \\
Claude-3-opus                 & 1867 & 254 & 8 & 87.59     & 94.84              & 76.49               \\
Qwen2.5-72b-Inst                 & 2710 & 378 & 5 & 79.92     & 91.02              & 78.54               \\ \bottomrule
\end{tabular}%
\end{table}

\begin{figure}
    \centering
    \begin{subfigure}[b]{0.45\linewidth}
        \centering
        \includegraphics[width=\linewidth]{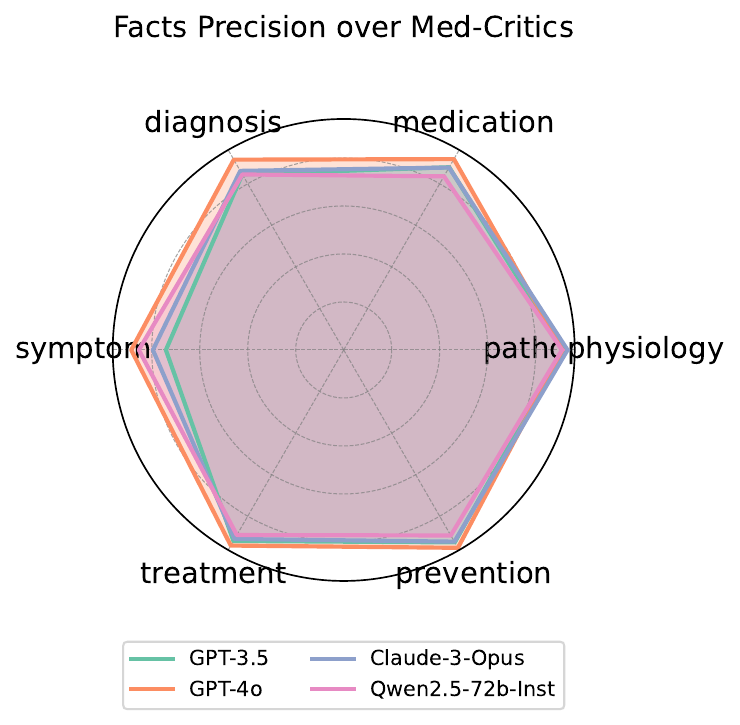}
        \caption{Verified as support ratio on multiple dimensions}
        \label{fig:model_comparison_radar}
    \end{subfigure}
    \hfill
    \begin{subfigure}[b]{0.45\linewidth}
        \centering
        \includegraphics[width=\linewidth]{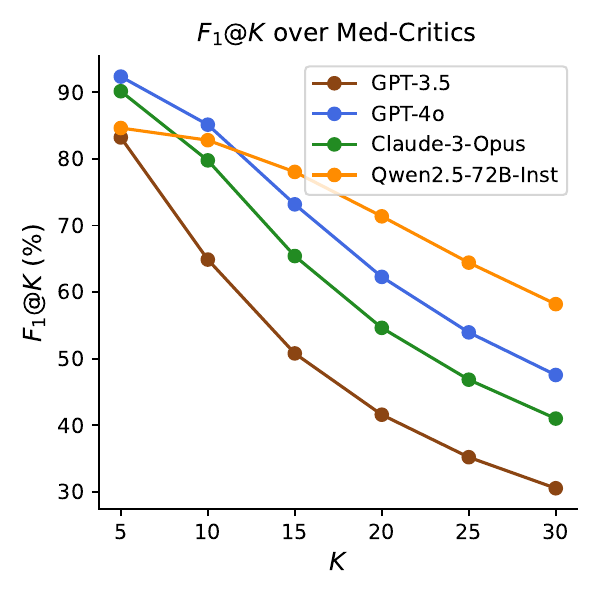}
        \caption{Response $F1@K$ metrics on different models}
        \label{fig:f1onK}
    \end{subfigure}
    \caption{Model performance on long-form medical text}
    \label{fig:combined}
\end{figure}

\paragraph{LLMs performance on long-form medical texts.}
While prior research has attempted to benchmark long-form factuality across LLMs \cite{minFActScoreFinegrainedAtomic2023,weiLongformFactualityLarge2024}, our works presents a more-comprehensive evaluation suitcase, and fine-grained medical dimensions of medical texts as show in Table~\ref{tab:dataset-statistics}.
For these reasons, we benchmark across four model families--GPT-3, GPT-4o, Claude and Qwen models.
We evaluate each model on the same random subset of 240 prompts from the \ourbench~benchmark, and report the verification results statistics in Table~\ref{tab:models-performance-comparsion}.
The results show that GPT-4o and Claude tend to say more and with with higher precision.

As shown in Figure~\ref{fig:f1onK}, the $F_{1}@K$ metrics of the GPT-4o on the \ourbench~benchmark are consistently outperform the other models.
The radar chart in Figure~\ref{fig:model_comparison_radar} provides a visual comparison of the models' performance across different dimensions of medical text.
The results indicate that GPT-4o exhibits the highest precision and recall, followed by Claude, Qwen, and GPT-3.5, especially in symptom, diagnosis dimensions.

\section{Discussion and Future Work}

In this paper, we tackled the challenge of verifying complex, long-form medical texts that often contain implicit claims. 
We proposed \oursfull, a framework that extracts self-contained facts from these texts and leverages retrieved information to perform adaptive, in-depth analyses starting from the initial claims. 
To evaluate its effectiveness, we curated \ourbench, a fine-grained dataset designed to assess factual verification by falsifying individual facts in long-form medical texts.
Our experiments on \ourbench~show that \ours~outperforms existing methods in factual verification tasks. 
This improvement stems from its ability to extract medical facts from implicit claims and its detailed, mechanism-level reasoning process, structured as a tree that consolidates retrieved knowledge. 
Looking ahead, we aim to further refine the claim extraction and retrieval components to enhance the accuracy and coverage of verifying complex medical texts. 
Additionally, by traversing the reasoning tree, \ours~can generate long-form chains of thought, enabling comprehensive responses to medical problems and demonstrating a deep understanding of underlying reasoning processes.

\bibliography{ref}
\bibliographystyle{plainnat}

\end{document}